\documentclass[journal]{IEEEtran}
\usepackage{amssymb}
\usepackage{amsmath} 
\usepackage[numbers,sort&compress]{natbib}

\usepackage{graphicx}
\usepackage[export]{adjustbox}
\usepackage{stfloats}
\usepackage{circledsteps} 
\usepackage{float}
\usepackage{booktabs} 
\usepackage{array}    

\usepackage[colorlinks,linkcolor=blue,citecolor=blue]{hyperref}
\usepackage{hyperref}
\usepackage{url}
\usepackage{fontawesome5} 

\ifCLASSINFOpdf

\else

\fi

\begin{document}

\title{TransDex: Pre-training Visuo-Tactile Policy with Point Cloud Reconstruction for Dexterous Manipulation of Transparent Objects \thanks{This work has been submitted to the IEEE for possible publication. 
Copyright may be transferred without notice, after which this version may no longer be accessible.}}

\author{Fengguan~Li, Yifan~Ma, Chen~Qian, Wentao~Rao, Weiwei~Shang$^\dagger$\\
  $^\dagger$Corresponding Author\\
  University of Science and Technology of China\\[8pt]
\includegraphics[height=1.1em, valign=t, trim={0 0 0 0.1cm}, clip]{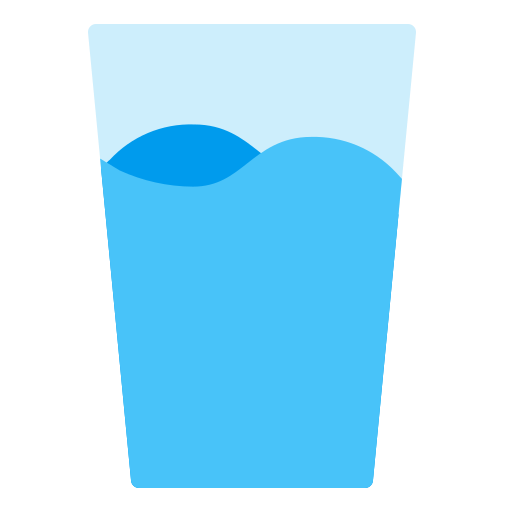}\textbf{Project page: }\href{https://TransDex.Github.io}{TransDex.Github.io} 
}

\maketitle

\begin{abstract}
Dexterous manipulation enables complex tasks but suffers from self-occlusion, severe depth noise, and depth information loss when manipulating transparent objects. 
To solve this problem, this paper proposes TransDex, a 3D visuo-tactile fusion motor policy based on point cloud reconstruction pre-training. 
Specifically, we first propose a self-supervised point cloud reconstruction pre-training approach based on Transformer. This method accurately recovers the 3D structure of objects from interactive point clouds of dexterous hands, even when random noise and large-scale masking are added. Building on this, TransDex is constructed in which perceptual encoding adopts a fine-grained hierarchical scheme and multi-round attention mechanisms adaptively fuse features of the robotic arm and dexterous hand to enable differentiated motion prediction. Results from transparent object manipulation experiments conducted on a real robotic system demonstrate that TransDex outperforms existing baseline methods. Further analysis validates the generalization capabilities of TransDex and the effectiveness of its individual components.
\end{abstract}
\def\abstractname{Note to Practitioners}
\begin{abstract}
 This work is motivated by the practical challenges faced by robotic systems when manipulating transparent objects, such as perceptual difficulties arising from depth information loss, severe noise, or self-occlusion. Existing methods often rely on intermediate perception completion steps, which are prone to failure in complex, dexterous manipulation scenarios. This limits their application in the real world. We propose a 3D visuo-tactile fusion motor policy that performs pre-training by reconstructing the 3D structure of objects from noisy, large-scale masked interactive point cloud data, and utilizes hierarchical encoding and multi-round attention mechanisms to adaptively fuse multimodal features, enabling differentiated motion prediction for robotic arm and dexterous hand. The proposed method is then validated on a real robotic system, demonstrating excellent performance and generalization capabilities. This work provides a practical, end-to-end solution for dexterous manipulation of transparent objects, reducing reliance on advanced depth sensors or complex optimization methods.
\end{abstract}
\begin{IEEEkeywords}
 Pre-Training, Visuo-Tactile Fusion, Dexterous Manipulation, Transparent Object Manipulation.
\end{IEEEkeywords}

%
\IEEEpeerreviewmaketitle

\section{Introduction}
%
%
%
%
\IEEEPARstart{D}{exterous} manipulation is a core capability that enables robots to perform tasks efficiently in real-world scenarios. Whether for routine tasks such as part assembly and stable grasping \cite{guzey2024see,chen2025dexforce,romero2024eyesight,liu2024masked,mao2024multimodal,guzey2023dexterity,yang2025equibot,wen2023any,zhao2025integrating}, or for complex manipulation like conical flask shaking and liquid pouring in automated laboratories \cite{li2025maniptrans,li2025chemistry3d}, accurate dexterous manipulation is essential. Achieving such tasks autonomously with multi-fingered dexterous hands is contingent on the development of robust and efficient learning policies.

In the field of learning-based robotic manipulation, researchers have successfully employed 2D visual inputs to train robots for a variety of tasks \cite{liu2024masked,mao2024multimodal,guzey2023dexterity,chi2025diffusion,chi2024universal}. With the development of this visual paradigm, recent research has placed growing emphasis on learning policies using 3D point cloud data \cite{Ze2024DP3,huang20243d,fu2025cordvip,li2025adaptive,yuan2024robot,ze2024generalizable,liu2024realdex,zhao2025robot}, which provides spatial structural information that better aligns with the physical properties of the real world. However, visual data is inadequate for comprehensive capture of interaction dynamics, especially in precision tasks that require contact feedback.

Advances in tactile sensing technology have led to a remarkable enhancement in the significance of tactile perception in dexterous manipulation. 
Tactile sensors, including Gelsight \cite{patel2020digger}, Uskin \cite{tomo2017covering} and PaXini \cite{Paxini2024TactileRobots}, provide essential data on contact force, pressure distribution and object pose, enabling compensation for limitations of vision in perceiving contact states. Consequently, visuo-tactile imitation learning has gradually gained traction \cite{guzey2024see,romero2024eyesight,liu2024masked,guzey2023dexterity,huang20243d,li2025adaptive,yuan2024robot,wu2025canonical}.

Despite the integration of multiple sensory modalities, dexterous manipulation in real-world scenarios still faces significant perceptual challenges. 
The multi-fingered structure of dexterous hands inevitably causes self-occlusion during interaction, which severely obscures critical spatial information about the hand and object postures \cite{fu2025cordvip}. 
When manipulating transparent objects, the perceptual insufficiency problem is exacerbated due to the effects of light refraction and reflection, resulting in noisy or even incomplete point cloud data \cite{jiang2023robotic}. 
Such perceptual ambiguity undermines the reliability of 3D representations and constrains the performance of motor policies that rely on 3D visual information. 
To address these limitations, existing research has predominantly focused on intermediate optimization steps such as depth completion or segmentation of transparent objects, followed by generating grasp poses based on refined information \cite{sajjan2020clear,dai2022domain,jiang2022a4t,li2023visual,bai2024cleardepth,zhai2025tcrnet}. Nevertheless, the accuracy of the intermediate optimization steps can possibly be affected by occlusion from the dexterous hands, potentially introducing and accumulating errors. 
It also struggles to adapt to real-time state changes during dynamic manipulation due to the open-loop grasping process. 
The development of end-to-end motor policies capable of dynamic dexterous manipulation in such perceptually ambiguous scenarios is still under-explored \cite{10597633,jiang2023robotic}.

As a result, the key challenges in dexterous manipulation of transparent objects can be summarized as follows:

1) Self-occlusion ambiguity: Multi-fingered structures cause inevitable self-occlusion during interaction, severely obscuring critical spatial information of hand-object postures.

2) Transparent object perception deficiency: Light refraction and reflection on transparent objects exacerbate insufficient perception, leading to noisy and missing point cloud data.

3) Intermediate optimization limitations: Existing methods rely on intermediate steps, whose accuracy is degraded by occlusion and fail to adapt to real-time dynamic state changes.

\begin{figure*}[!htb] 
		\centering 
		\includegraphics[width=0.95\textwidth]{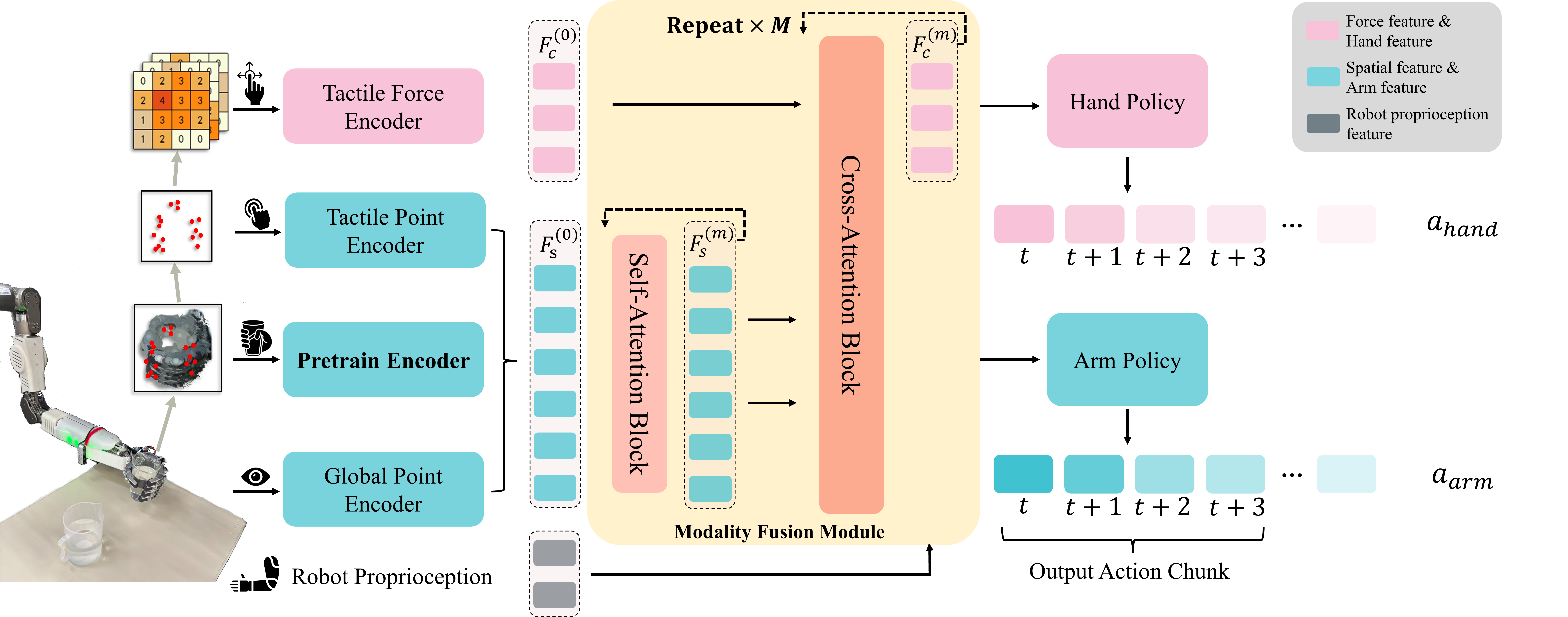}
		\caption{\textbf{The Overall Framework of the Pre-Training-Based Visual-Tactile Fusion Motor Policy, TransDex.} Perceptual information enters the encoder through a fine-grained, hierarchical manner. The hand-object interaction point cloud utilizes a pre-trained encoder, followed by an attention fusion module and two policy heads to achieve feature integration and differentiated action prediction. } 
		\label{Fig1} 
\end{figure*}
Based on previous observations, this study proposes TransDex, a policy that enhances manipulation performance through robust 3D object perception. The objective of this study is to develop a robust end-to-end visuo-tactile motor policy that directly addresses perceptual ambiguities, such as occlusion and interference of transparent objects, without relying on independent perception modules. The overall framework of TransDex is illustrated in Fig.\ref{Fig1}. Specifically, we first design a Transformer-based point cloud reconstruction pre-training framework. 
The pre-trained network reconstructs complete geometric structures of objects from randomly masked and noisy interactive point clouds of dexterous hands, thereby enhancing the understanding of object shape and pose under sparse observations. Building upon this foundation, we further propose a visuo-tactile motor policy. The policy employs a fine-grained decomposition of multi-modal perceptual information, followed by hierarchical encodings. Specifically, interactive point clouds of dexterous hands are processed by the pre-trained encoder. Through multiple self-attention and cross-attention modules for computation, the policy adaptively fuse features of the robotic arm and dexterous hand to enable differentiated motion prediction, based on distinct combinations of information.

We performed transparent object manipulation experiments on a real robotic system, including liquid pouring, flask shaking, and cup rotating. The comparative results demonstrate that TransDex outperforms baseline policies in terms of success rate under such a scenario where perceptual information is severely ambiguous. We also evaluated the performance of the pre-trained network on the point cloud reconstruction task. Ablation studies verified the effectiveness of both the pre-training method and each component of TransDex. Further analysis reveals that TransDex exhibits strong generalization capability and robustness against unseen and complex scenarios.

In summary, the main contributions of this work are as follows:

1) We propose a Transformer-based point cloud reconstruction pre-training method that effectively enhances the robot's understanding of object shape and pose in perceptually ambiguous scenarios.

2) Building on the pre-training method above, we propose an end-to-end 3D visuo-tactile motor policy. It refines the decomposition and encoding of perceptual information, enabling differentiated action prediction through feature combination, without relying on independent perception modules.

3) We verified the effectiveness and robustness of TransDex via a series of experiments on transparent object reconstruction and manipulation conducted on a real robotic system.

The rest of this article is organized as follows. Section \ref{sec:related_works} reviews related works. Section \ref{sec:transdex_policy} details the proposed end-to-end visuo-tactile motor policy, TransDex. Section \ref{sec:experiments} presents experimental validations of TransDex. Finally, Section \ref{sec:conclusion} concludes  this work.
\begin{figure*}[ht] 
		\centering 
		\includegraphics[width=0.9\textwidth]{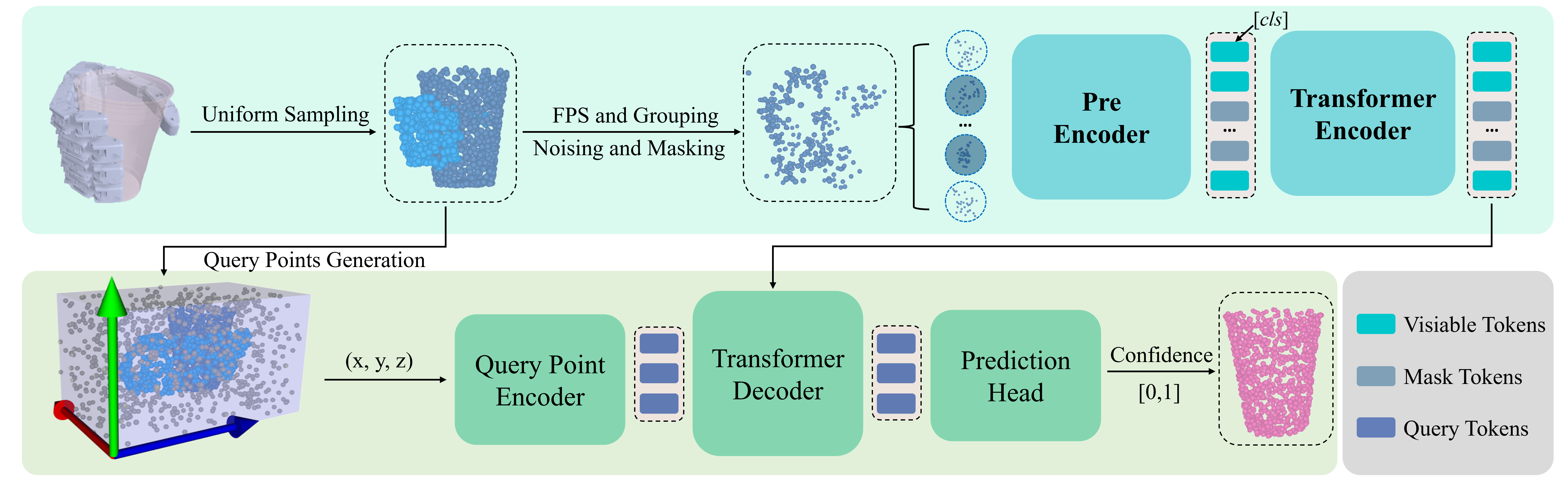}
		\caption{\textbf{Pre-Training Framework.} First, noise is added to and masked from the original hand-object interaction point cloud to generate input data. Then, features are extracted using a pre-encoder and a Transformer encoder. Finally, the generated query points and a Transformer decoder are employed to accomplish the point cloud reconstruction task for the object.} 
		\label{Fig2} 
\end{figure*}
\section{Related Work}
\label{sec:related_works}

\subsection{Visuo-Tactile Fusion}
As tactile sensation can capture subtle contact information that vision cannot, numerous studies have focused on achieving robotic perception and manipulation through the fusion of vision and tactile sensing. A common fusion approach involves encoding the two modalities separately, followed by the fusion or concatenation of the resulting embedded features \cite{guzey2024see,romero2024eyesight,liu2024masked,sun2025vtao}. Yuan et al. \cite{yuan2024robot} proposed ``robot synesthesia'', which unified visual and tactile data into point cloud input by calculating the spatial coordinates of contact points via forward kinematics. 3D-ViTac was built based on this idea and employed a higher tactile resolution \cite{huang20243d}. However, both methods primarily utilize only normal force information, and lack effective feature-level integration. Moreover, researchers also proposed robot-agnostic visuo-tactile perception models to reconstruct hand-object interaction shapes \cite{xu2023visual,jiang2024capturing}, but these methods have not yet been extended to robot motion generation and control. Overall, existing visuo-tactile fusion methods still exhibit limitations in the fine-grained decomposition of perceptual information and guidance for dynamic manipulation.

\subsection{Pre-training for Robotics}
In the field of robotic pre-training, unsupervised learning is a prevalent paradigm. Pre-training methods for visuo-tactile modalities can be categorized into two broad types: one involves aligning the feature spaces of visual and tactile modalities during pre-training \cite{liu2024masked,sun2025vtao,zhu2025touch,wang2024scaling,gano2024low}, while the other focuses on pre-training within data from a single modality \cite{guzey2023dexterity,wu2025canonical,nair2023r3m,yan2024dnact,ze2023h,yuan2025learning}. Common pre-training enables encoders to develop the ability to distinguish between different data samples or accomplish specific sub-tasks, thereby providing effective guidance for training downstream policies. For instance, researchers guide the tactile encoder during pre-training to focus on learning force-related features in 3DTacDex \cite{wu2025canonical}, while Zhu et al. \cite{zhu2025touch} use encoders to reconstruct tactile images from visual and tactile inputs, thereby enhancing the model's ability to infer local contact conditions and understand cross-modal correlations. However, extant pre-training methods frequently neglect to leverage the 3D structural information of objects during interaction and lack explicit modeling of real-world disturbances such as occlusion and noise, which has a consequence of limiting their capabilities in complex scenarios.

\subsection{Robotic Manipulation for Transparent Objects}
The complex refraction and reflection effects of transparent objects present significant challenges for target perception and state recognition in motor policy, particularly in 3D motor policy. Existing solutions for robotic manipulation of transparent objects typically use a two-stage process: first, perceptual completion via depth estimation or 3D reconstruction; then, grasp pose generation using the completed results. Representative models following this paradigm include ClearGrasp \cite{sajjan2020clear}, SwinDRNet \cite{dai2022domain}, A4T \cite{jiang2022a4t}, and ClearDepth \cite{bai2024cleardepth}. Nevertheless, the occlusion caused by dexterous hands may affect the completion effect and the inaccuracy could be amplified during the subsequent phase, directly leading to grasp failures. Furthermore, these methods are typically limited to the execution of grasping or rely on pre-programmed motion sequences to complete subsequent manipulation. Therefore, real-time end-to-end manipulation policies for transparent objects hold significant research value. In Maniptrans \cite{li2025maniptrans}, researchers proposed a method for transferring human skills to dexterous robotic hands and conducted evaluations of imitation learning, including manipulation of transparent objects. However, its highest reported success rate was only 18.44\%, indicating limited practical utility.
\section{Method}
\label{sec:transdex_policy}
The overall framework  operates in two phases: 1) The pre-training framework encodes randomly masked and noisy hand-object interaction point cloud, and reconstructs the complete object point cloud via a decoder, as described in \ref{subsec:Pre-training Based on Point Cloud Reconstruction}. This process endows the pre-trained network with robust object comprehension against incomplete perceptual scenarios. 2) Building upon the pre-trained network, the motor policy accomplishes the extraction and multi-modal integration of fine-grained visuo-tactile information, ultimately generating differentiated motion predictions for both the robotic arm and the dexterous hand, detailed in \ref{subsec:Visuo-Tactile Fusion Motor Policy}.
\subsection{Pre-training Based on Point Cloud Reconstruction}
\label{subsec:Pre-training Based on Point Cloud Reconstruction}

As illustrated in Fig.\ref{Fig2}, the pre-training task is defined as follows: for the original hand-object interaction point cloud, under the interference of random masking and added noise, an encoder-decoder structure is utilized to predict the probability that densely generated query points in 3D space belong to the target object. This is formally represented by the following equation:
\begin{equation}
    \mathbb{P}(q \in \mathcal{O} \mid \tilde{P}_{o}') = f_\theta(\tilde{P}_{o}'; q)
    \label{eq:placeholder_label}
\end{equation}
where $q$ denotes query points, $\mathcal{O}$ represents target object point cloud, ${\widetilde{P}}_{o}'$ signifies the point cloud under perceptual interference, and $f_\theta$ represents the pre-trained network.

The pre-training framework is comprised of several key components:

\subsubsection{Pre-training Dataset} First, a dexterous grasping scene is constructed on PyBullet \cite{coumans2016pybullet}. The objects in the scene are primarily sourced from the ClearPose dataset \cite{chen2022clearpose}, encompassing common daily items and laboratory instruments such as various types of cups, vases, flasks, etc. Subsequently, points are uniformly sampled on the dexterous hand and the object respectively to generate a complete hand-object interaction point cloud:
\begin{equation}
    P_o^t = P_{hand}^t + P_{object}^t \in \mathbb{R}^{n \times 3}
    \label{eq:placeholder_label}
\end{equation}

Subsequently, random noise is added to the object point cloud, resulting in a noisy point cloud ${\widetilde{P}}_{o}\in \mathbb{R}^{n \times 3}$. Key parameters, including object size, position, as well as the direction, percentage, and magnitude of the added noise, are varied to ensure that the pre-trained network adapts to various noisy scenarios. Note that only dexterous hand is retained in the scenario to guide the model to focus on learning to reconstruct the object point cloud from the grasping configuration of the dexterous hand.

\subsubsection{Grouping and Random Masking} Given ${\widetilde{P}}_{o}$, $c$ center points $\left\{C_i \right\}_{i=1}^c$ were selected using Farthest Point Sampling(FPS). The points surrounding each center are then aggregated into patches via the K-Nearest Neighbors(KNN) algorithm. These patches are randomly masked at a ratio of $R_{mask}$, resulting in masked point cloud ${\widetilde{P}}_{o}'$. By randomly masking a portion of the patches, the model is forced to learn global-local feature relationships, thereby enhancing its generalization capability when dealing with incomplete perceptual data.

\subsubsection{Point Cloud Feature Extraction} Visible patches are fed into a pre-encoder to convert them into Transformer-compatible tokens. Additionally, a learnable $cls$ token is introduced to aggregate features from all patches within the subsequent Transformer encoder and aid in distinguishing between different samples. The resulting visible patch tokens are then processed by the Transformer encoder to further extract complex global correlations and local detail features from hand-object interaction point cloud. The final output is a set of high-dimensional latent features, including the feature of $cls$ token ${F}_{[cls]}$.

\subsubsection{Query Point Generation and Decoder} The objective of the pre-training is to enable the aforementioned encoder to interpret object shapes from the noisy and masked hand-object interaction point cloud. This is achieved by performing a query point classification task.  We first generate dense random points in the 3D space and merge them with the original point cloud as query points. Then the complete point cloud of the object $P_{object}^t$ is labeled as positive query points $\left\{ Q_{p_i}^{t} \right\}_{i=1}^{N_{t}} \in \mathbb{R}^{N_{t} \times 3}$, while the remaining points in the space (including the point cloud of the dexterous hand $P_{hand}^t$) are treated as negative query points $\left\{ Q_{p_i}^n \right\}_{i=1}^{N_{n}} \in \mathbb{R}^{N_{n} \times 3}$. Specifically, if the minimum distance from a negative point to any positive point is smaller than the minimum distance between any two positive points, the negative point is reclassified as positive. This is because such points are sufficiently close to the object's surface to be considered part of the object. It is important to note that positive query points $\left\{ Q_{p_i}^{t} \right\}_{i=1}^{N_{t}}$ are not simply a normalized point cloud of the object. Rather, they are uniformly sampled based on the actual pose of the object within the original point cloud coordinate system of each specific grasping instance. This design forces the model to learn how to interpret the shape and pose of the object simultaneously from sparse and noisy point clouds.

The overall query points $\left\{ Q_{p_i} \right\}_{i=1}^{N}$, where $N=N_{t}+N_{n}$, are processed by the query point encoder to generate positional embeddings. These are then fed into the Transformer decoder along with the latent features output by the encoder. The decoder produces a latent representation for each query point, which is then passed through a binary prediction head to output a confidence score $\left\{ p_{i} \right\}_{i=1}^{N}$ for each query point belonging to the target object. Critically, the encoder-decoder architecture is designed to prevent information leakage from masked portions of the original point cloud, including the positions or embeddings of masked patches. 
This constraint mirrors real-world conditions in which the robot has no access to information about regions that are not in view. 
Enforcing this design compels the model to genuinely learn how to efficiently utilize visible information and robustly infer the structure of the object. 
This ensures the development of reliable object perception capabilities for real-world deployment.

The training objective of the pre-trained network is to minimize the classification loss of the query points and the similarity of  $cls$ features between different samples:
\begin{equation}
\begin{aligned}
    \min_{\theta} \mathbb{E}_{(\tilde{P}_{o}',P_{object}^t )} & \left[ \mathcal{L}_{cls} \left( p, y(Q_p, P_{object}^t ) \right) \right] \\ 
    & + \lambda \mathbb{E}_{\tilde{P}_{o}'} \left[ \mathcal{L}_{contrast} \left( F_{[cls]} \right) \right]
\end{aligned}
\label{eq:placeholder_label}
\end{equation}
where $\theta$ denotes the network parameters, $y$ denotes the true binary label of the query point, $\lambda$ is the weighting coefficient, and $\mathcal{L}_{contrast}$ denotes the contrast loss.

\subsection{Visuo-Tactile Fusion Motor Policy}
\label{subsec:Visuo-Tactile Fusion Motor Policy}

After completing the pre-training based on point cloud reconstruction, only the encoder of the pre-trained network is retained, while the decoder is discarded. Endowed with pre-trained weights, the encoder can comprehend object shape and pose from perceptual ambiguous inputs, thereby providing the perceptual features essential for training the downstream motor policy. As shown in Fig.\ref{Fig1}, the visuo-tactile motor policy comprises four stages:

\subsubsection{Point Cloud Processing} The two point clouds obtained from the depth camera are first unified into the robot's base coordinate system. Fine registration is then performed using the Iterative Closest Point (ICP) algorithm to mitigate the impact of hand-eye calibration errors on merging the point clouds. The registered point clouds are then cropped and standardized to a fixed value $p$ using Farthest Point Sampling (FPS), resulting in a global point cloud $P_g\in\mathbb{R}^{p\times6}$. Here, the six channels represent three-dimensional coordinates and three RGB color channels. Subsequently, an artificial cropping volume is defined within the base coordinate frame of the dexterous hand. The global point cloud is then cropped within this volume, and the color dimensions are discarded, resulting in the vision-based hand-object interaction point cloud $P_{h_o}^v$.

Dense tactile sensing can complement sparse visual point clouds by providing detailed information about object boundaries, which is valuable for transparent objects. Therefore, the 3D position of each taxel on the tactile sensor that registers a force reading is computed via forward kinematics using real-time robotic joint angles $q\in\mathbb{R}^{23}$, expressed in the robot's base coordinate frame as $P_{h_o}^t$. These tactile-based points are then incorporated into $P_{h_o}^v$, resulting in the final hand-object interaction point cloud $P_{h_o}=P_{h_o}^t+P_{h_o}^v$.

\subsubsection{Perceptual Encoding} Using independent encoders for different modalities, it captures global semantic information while progressively mining fine-grained local details. This provides precise feature support for subsequent multimodal fusion and motion generation policy. The specific implementation is as follows:

First, for the global point cloud $P_g$, it is encoded using a global point cloud encoder composed of MLP layers, LayerNorm layers, and a maximum pooling layer, which ultimately produces the perceptual feature $F_{g}$.

Second, for the hand-object interaction point cloud $P_{h_o}$, the pre-trained encoder is utilized for feature extraction. The pre-trained encoder outputs multiple feature tokens, from which only the $F_{[cls]}$ corresponding to the $cls$ token is used here. To better adapt the feature for the downstream motor policy, a learnable MLP is appended after $F_{[cls]}$, resulting in $F_{[cls]}'$ as the final perceptual feature of $P_{h_o}$.

Third, in addition to supplementing boundary information in the hand-object interaction point cloud, tactile-based points can also assist in providing contact pose references for interaction force adjustment and contact position optimization. Therefore, tactile-based points $P_{h_o}^t$ is encoded to yield $F_{pos_t}$.

Finally, to efficiently process the array-based 3D force signals from the tactile sensors, they are first reshaped into an image format $I_f$, and then processed by a CNN-based tactile encoder, resulting in the perceptual feature $F_{t}$ for the tactile force information.

\subsubsection{Modality Fusion Module} The feature fusion module consists of multiple self-attention and cross-attention blocks. The global point cloud feature $F_{{g}}$, the hand-object interaction point cloud feature $F_{[cls]}'$, and the tactile-based point feature $F_{{pos}_t}$ form a progressively refined spatial relationship. These three core spatially relevant features are fed into a self-attention module to achieve feature aggregation, learning the association patterns and key information within the pan-visual modalities. This process yields the self-attention processed latent feature $F_{{s}}^{(1)}$. $F_{{s}}^{(1)}$ subsequently serves as the key and value, while the tactile force feature $F_{{t}}$ serves as the query, these are then input into a cross-attention module for deep interaction. Through attention weight allocation, the tactile force information modulates the features from other modalities in a targeted manner, resulting in the latent feature $F_{{c}}^{(1)}$. This computational process is repeated for $M$ iterations, ultimately yielding the self-attention output of and the cross-attention output of the $M^{{th}}$ round, denoted as $F_{{s}}^{(M)}$ and $F_{{c}}^{(M)}$, providing differentiated multimodal fused representations for downstream motion control tasks. This process is formally represented as follows:
\begin{equation}
\begin{aligned}
        \begin{cases}
        F_{{s}}^{(m)} = \mathrm{SelfAtten}\left(F_{{s}}^{(m-1)}\right) \\
        F_{{c}}^{(m)} = \mathrm{CrossAtten}\left(F_{{c}}^{(m-1)}, F_{{s}}^{(m)}\right)
        \end{cases}
\end{aligned}
\label{eq:placeholder_label}
\end{equation}
where $m = 0,1, \dots, M $, $F_{{s}}^{(0)} = \left[ F_{g}, F_{[cls]}', F_{pos_t} \right] $, $ F_{c}^{(0)} =\left[ F_{t}\right] $.

\subsubsection{Feature Combination and Task Adaptation} The self-attention output $F_{s}^{(M)}$, while lacking direct force information involvement, integrates global point cloud data with local hand-object details, augmented by tactile-based points data. This results in a feature representation encompassing both environmental awareness and interactive understanding, making it siutable for guiding the pose trajectory and spatial offset correction of the robotic arm. In contrast, the cross-attention output $F_{c}^{(M)}$ incorporates cross-modal modulated features that are primarily influenced by tactile force feedback. It emphasises details such as force feedback and contact states during tactile interaction and is augmented with visual information to perceive object shape and pose. Consequently, it is well-suited to the fine-grained regulation of dexterous hand motion, such as grasp pose adjustment and interactive force control. Therefore, $F_{s}^{(M)}$ is used to guide the action generation for the robotic arm, while $F_{c}^{(M)}$ is used to guide the dexterous hand.

\subsubsection{Action Generation} For the lower-dimensional robotic arm end-effector pose $a_{a}\in\mathbb{R}^{pred_T\times6}$, a policy head ${\textbf{P}}_{a}$ composed of MLP layers is used for prediction with $F_{s}^{(M)}$ and the robot's proprioceptive state $s\in\mathbb{R}^{pred_T\times22}$, where $pred_T$ represents the timesteps for action prediction. For the high-dimensional dexterous hand actions $a_{h}\in\mathbb{R}^{pred_T\times16}$, a conditional denoising diffusion model \cite{chi2025diffusion,ho2020denoising} serves as the dexterous hand policy head ${\textbf{P}}_{h}$, which integrates $F_{c}^{(M)}$ and $s$ as conditions. It iteratively denoises a random Gaussian noise vector ${a_{h}}^K$ over $K$ steps to obtain the target action ${a_{h}}^0$. The specific action generation pipeline is as follows:

The state input for imitation learning, denoted as $S_{in}$, is defined as an observation sequence from the past $obs_T$ timesteps:
\begin{equation}
\begin{aligned}
S_{{in,T}} =& <\{P_g, P_{h_o}, P_{h_o}^t, I_f, s \}_{T-{obs}_T+1},  \\ 
& \dots,  \{P_g, P_{h_o}, P_{h_o}^t, I_f, s \}_T>
\end{aligned}
\label{eq:placeholder_label}
\end{equation}

Perceptual encoding and multi-modal feature fusion are performed based on the state input as follows:
\begin{equation}
\begin{aligned}
\left\{  F_{{s}}^{(M)},  F_{{c}}^{(M)} \right\} = \mathrm{Fusion}\left(\mathrm{Encoder}\left( S_{{in}} \right) \right)
\end{aligned}
\label{eq:placeholder_label}
\end{equation}

Actions for the robotic arm and the dexterous hand are predicted based on the fused features as follows:
\begin{equation}
{\widehat{a}_{a}}={\textbf{P}}_{a}\left[F_{s}^{\left(M\right)}\right]
\label{eq:placeholder_label}
\end{equation}
\begin{equation}
\begin{aligned}
{a_{h}}^{k-1}=&\alpha_k\left({a_{h}}^{k}-\gamma_k\epsilon_\theta\left({a_{h}}^{k},k,F_{c}^{(M)}\right)\right)\\
&+\sigma_k\mathcal{N}(0,I)
\label{eq:placeholder_label}
\end{aligned}
\end{equation}
where $\alpha_k$, $\gamma_k$ and $\sigma_k$ are parameters determined by the noise scheduler and the denoising step $k$, $\epsilon_\theta$ represents the denoising network, and $\mathcal{N}\left(0,I\right)$ represents Gaussian noise.

The loss function is calculated as
\begin{equation}
\mathcal{L}_{a}=\mathrm{MSE}({\widehat{a}_{a}},\ {a_{a}})
\label{eq:placeholder_label}
\end{equation}
\begin{equation}
\mathcal{L}_{h}=\mathrm{MSE}(\epsilon^k,\epsilon_\theta(\bar{\alpha}_k{a_{h}}^{0}+\bar{\beta_k}\epsilon^k,k,F_{c}^{(M)}))
\label{eq:placeholder_label}
\end{equation}
\begin{equation}
\mathcal{L}_{t}=\lambda_1\mathcal{L}_{a}+\lambda_2\mathcal{L}_{h}
\label{eq:placeholder_label}
\end{equation}
where $\epsilon^k$ denotes the noise at step $k$ in the diffusion process; $\bar{\alpha}_k$ and $\bar{\beta}_k$ are noise scheduling parameters; $\lambda_1$ and $\lambda_2$ are loss weighting factors. 
\section{Experiments}
\label{sec:experiments}
We performed transparent object manipulation experiments on a physical robotic platform to validate the following: 1) the applicability of TransDex to transparent object manipulation tasks in real-world scenarios, and its performance advantages over existing baseline methods; 2) the effectiveness of the pre-trained network in point cloud reconstruction under perceptually ambiguous scenarios in both simulation and reality, and its contribution to the performance of TransDex; 3) the impact of other components of TransDex on manipulation performance; 4) the robustness of TransDex when confronted with unseen objects or other perceptual disturbances.
\subsection{Robotic System Setup}

\begin{figure}[t] 
		\centering 
		\includegraphics[width=0.48\textwidth]{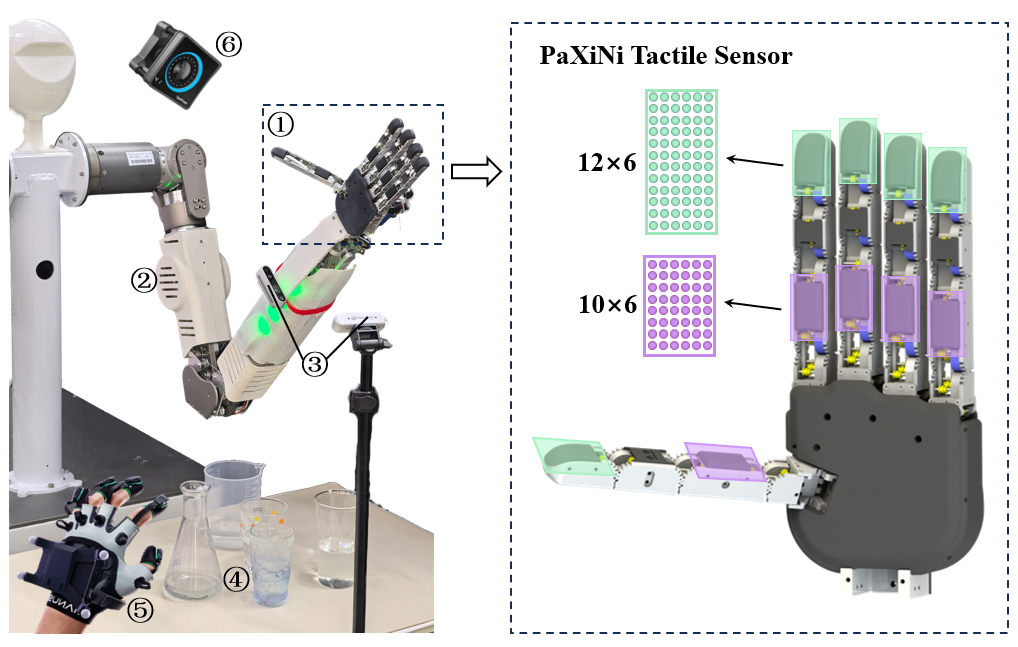}
		\caption{\textbf{Robotic System Setup:} \Circled{1} a 16-DOF dexterous hand, with array tactile sensors equipped on the fingertips and finger pads; \Circled{2} a 7-DOF robotic arm; \Circled{3} depth cameras; \Circled{4} experimental items; \Circled{5} a data glove; \Circled{6} a motion capture camera.} 
		\label{Fig3} 
\end{figure}
As shown in Fig.\ref{Fig3}, the real robotic system comprises a 7-DOF humanoid robotic arm and a 16-DOF dexterous hand \cite{ma2025development}. PaXini tactile array sensors are installed on the fingertips and fingerpads of each finger of the dexterous hand. The fingertip sensor array has a configuration of 12×6, while the fingerpad array is 10×6. Each tactile unit can measure two-dimensional tangential forces and one-dimensional normal force. Two RealSense D435 cameras are positioned at the wrist of the robotic arm and around the workbench respectively. The robot is teleoperated using a Manus-Metaglove in combination with an OptiTrack motion capture system to collect expert demonstration data. 
With its sophisticated perceptual capabilities enabled by multiple sensors and the fine manipulation capabilities of a high-DOF robot, this system is well-suited for manipulating transparent objects.
\subsection{Description of Dexterous Tasks}
To evaluate the performance of TransDex, our experiment focused on dexterous manipulation tasks of transparent objects. As shown in Fig.\ref{Fig4}, the following tasks were designed:

\subsubsection{Pouring water} The robot must grasp a glass cup containing water from the table, move it above a measuring cup, rotate its wrist to orient the opening downward and pour the water into the measuring cup. The task is considered successful if the water is poured into the measuring cup and the glass cup is grasped stably without slipping and dropping during the process.

\subsubsection{Shaking flask} The robot must move next to a conical flask, grip it with its thumb and index finger and use its remaining fingers to shake it back and forth. The flask contains a mixture of glucose, sodium hydroxide (NaOH) and methylene blue solution. Initially transparent, the solution turns purple after sufficient shaking. The task is considered successful if the robot can shake the flask continuously for over half a minute, causing the solution to change to purple, without dropping it. 
Some baseline or ablation policies were observed to tend to press the three fingers tightly against the flask, resulting in small-amplitude shaking that did not meet the intent of the task. If this shaking pattern causes discolouration, it is counted as a 0.5 success.

\subsubsection{Rotating Glass Cup} The robot first moves near the cup, then coordinates its thumb, index, and middle fingers to continuously rotate the cup. During rotation, the thumb contacts the cup and pushes it outwards, while simultaneously the other two fingers rotate the cup towards the palm, completing one rotation together. The fingers subsequently release and re-establish contact with the cup, repeating the process. The task is successful if the cup remains standing and the cumulative rotation angle exceeds 90 degrees.
\subsection{Baselines}
To validate the effectiveness of TransDex, a comparative evaluation of its manipulation task performance was conducted against the following three baseline methods based on 3D point clouds:

\subsubsection{3D-ViTac \cite{huang20243d}} A robot synesthesia learning approach that integrates tactile-based points and force information into a point cloud representation for joint training. The original method utilizes only one-dimensional normal force information. In our experiments, it was extended to utilize three-dimensional force data to obtain richer force information.

\subsubsection{DP3 \cite{Ze2024DP3}} A simple yet effective 3D point cloud imitation learning method. The original model architecture was used in the experiments, without incorporating the tactile modality.

\subsubsection{3DTacDex-P \cite{wu2025canonical}} A method that employs a novel representation scheme for array-style tactile data and models tactile information using a graph structure. In our experiments, the RGB image input in the model was replaced with point cloud input consistent with our policy.

These three baseline methods encompass visual point cloud driven policy, visuo-tactile point cloud fusion, and novel tactile representation schemes, facilitating a comprehensive evaluation of the advantages of TransDex across key technical dimensions of dexterous manipulation.
\begin{figure*}[t] 
		\centering 
		\includegraphics[width=0.98\textwidth]{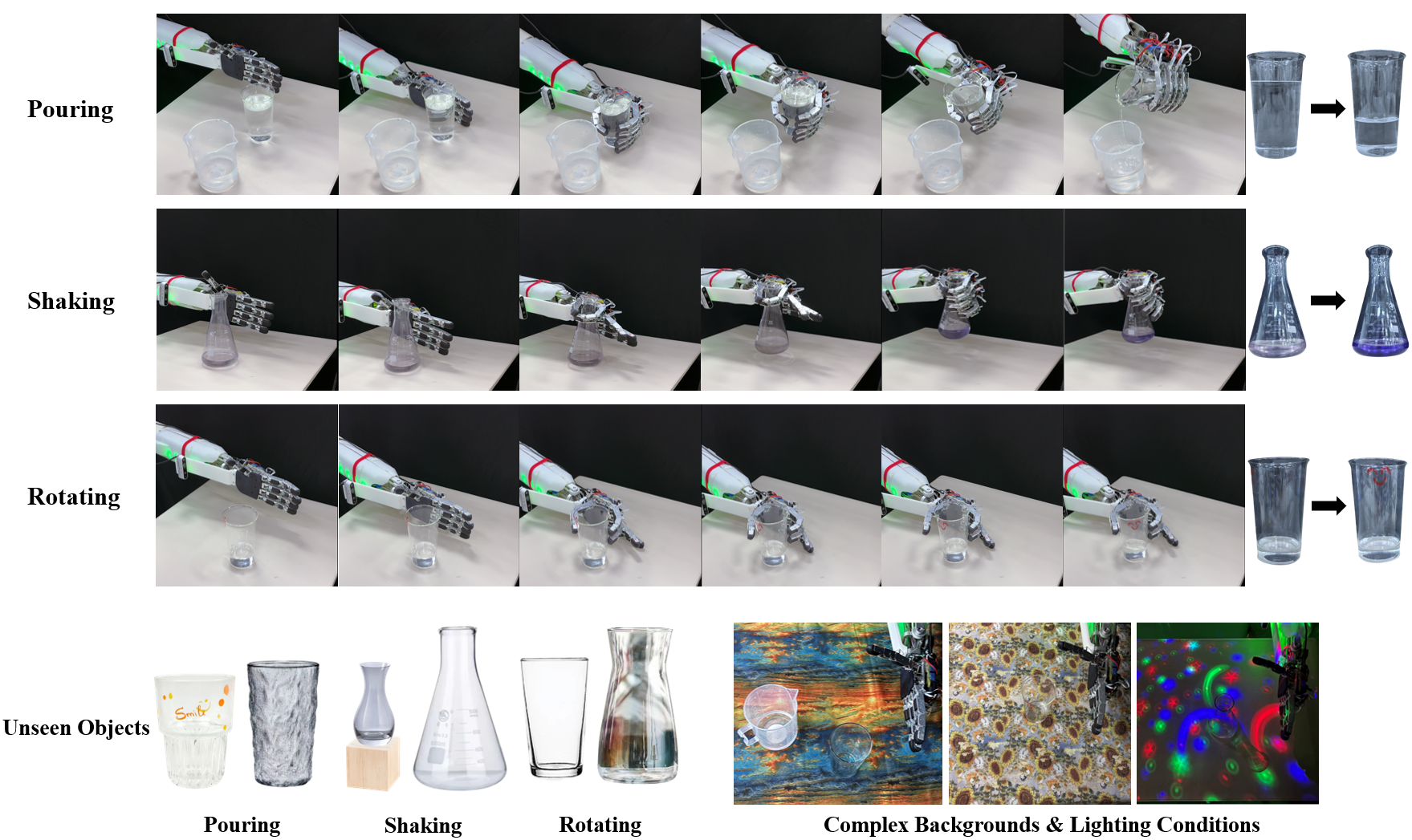}
		\caption{\textbf{Visualization of Policy’s rollout on Three Transparent Object Manipulation Tasks}, including pouring, shaking, and rotating. The unseen objects used in the test, as well as the complex backgrounds and lighting conditions, are shown at the bottom.} 
		\label{Fig4} 
\end{figure*}
\subsection{Effectiveness Comparison of Manipulation Policies}
\begin{table}[t]
  \caption{Success Rate of TransDex and Baseline Policies}
  \label{table1}
    \centering
  \begin{tabular}{l c c c c}
    \toprule
    \textbf{Method} & \textbf{Pouring} & \textbf{Shaking} & \textbf{Rotating} & \textbf{Avg} \\
    \midrule
    3D-ViTac        & 0\%              & 0\%              & 0\%              & 0\%       \\
    DP3             & 10\%             & 0\%              & 50\%             & 20\%      \\
    3DTacDex-P     & 70\%             & 50\%             & 20\%             & 47\%      \\
    \textbf{TransDex(Ours)} & \textbf{100\%}    & \textbf{80\%}    & \textbf{70\%}    & \textbf{83\%} \\
    \bottomrule
  \end{tabular}
  \vspace{0.7em}
  
  \footnotesize
  \raggedright
  
  \textit{\textbf{ *Note}}: Bolded content denotes the highest success rate of each task; same for the subsequent tables.
\end{table}
As shown in Table \ref{table1}, the proposed policy achieved an average success rate of 83\%, which is significantly higher than that of all other baseline methods across all tasks. Notably, all attempts with 3D-ViTac as the policy failed, as while the robot was capable of approaching the object effectively, it consistently failed to establish contact with or lift the object. DP3 also performed poorly across the three tasks, with an average success rate of only 20\%. During experiments, unstable end-effector movements with significant oscillations was observed, leading to task failure. This aligns with the findings of CordVip \cite{fu2025cordvip} and Rise \cite{wang2024rise}: point cloud-based policies generally fail to achieve high manipulation performance when the quality of point cloud data for transparent objects acquired by depth cameras is poor. Furthermore, the lack of tactile information in DP3 makes it particularly challenging to complete manipulation tasks on transparent objects using only incomplete visual data.

Among the baseline methods, the 3DTacDex-P policy performed best, achieving an average success rate of 47\%. While this policy performed well in the pouring task, it frequently exhibited significant end-effector jitter. In the shaking task, the robot's middle, ring, and little fingers typically remained pressed firmly against the flask after gripping it, resulting in insufficient shaking amplitude. For the rotating task, 3DTacDex-P demonstrated unstable performance, with difficulties in coordinating finger movements, resulting in a success rate of only 20\%. While the policy's well-designed representation endows it with excellent tactile perception capabilities, the inherent sparsity of the point cloud data ultimately limits its manipulation performance.

By contrast, owing to the integration of point cloud reconstruction pre-training, fine-grained feature encoding and feature combination-prediction design, TransDex can derive robust object representations from sparse point clouds. It exhibits strong robustness against incomplete point cloud information and demonstrates better arm-hand coordination during motion. Consequently, it achieved better performance across all three tasks.
\subsection{Effectiveness of Point Cloud Reconstruction Pre-training}

\begin{figure}[t] 
		\centering 
		\includegraphics[width=0.47\textwidth]{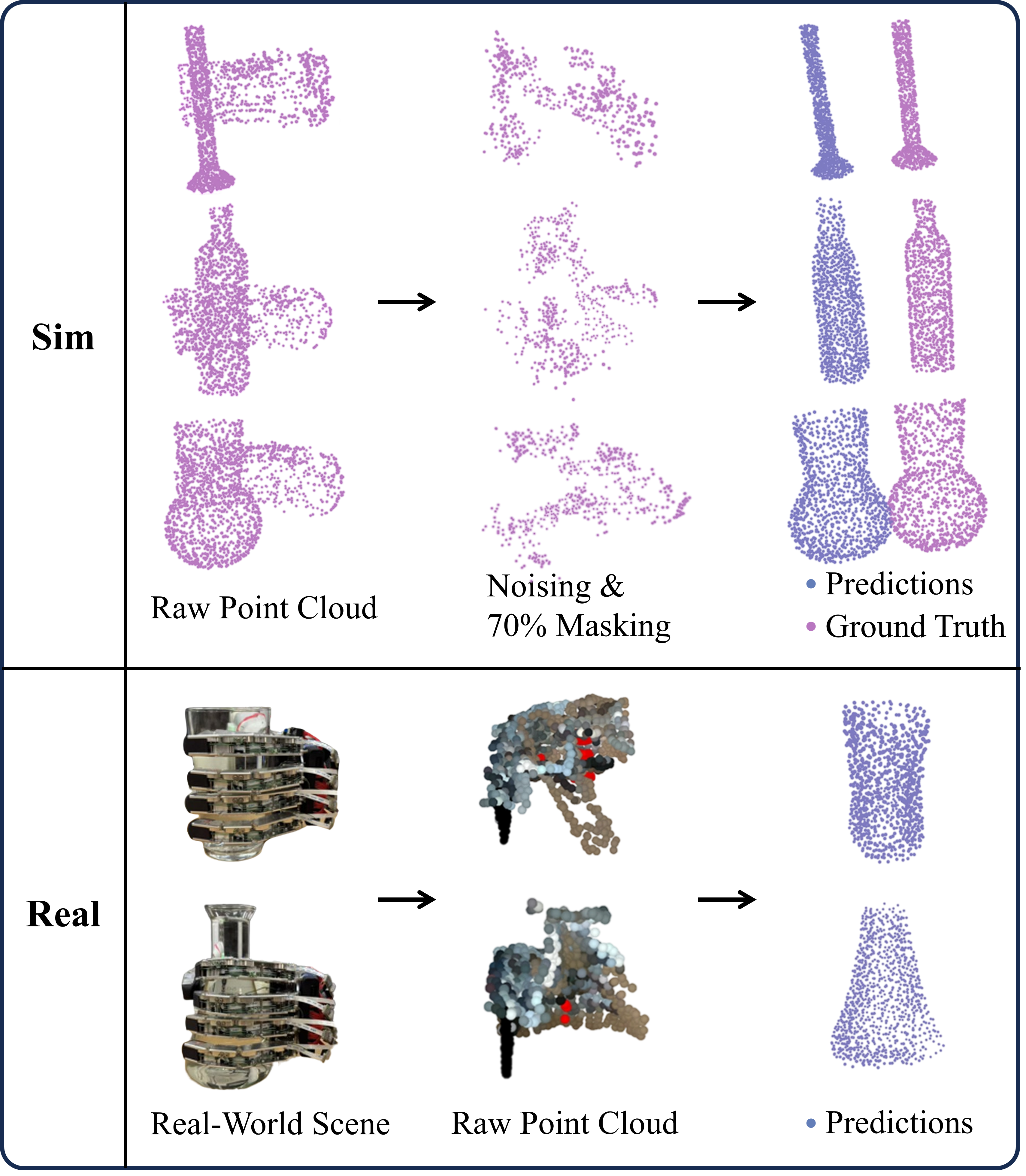}
		\caption{\textbf{Visualization of Point Cloud Reconstruction Performance in Pre-Training Tasks and Real-World Transfer Reconstruction Outcomes.}} 
		\label{Fig5} 
\end{figure}
Based on the pre-trained point cloud reconstruction network, the point cloud of the target object can be reconstructed by generating dense query points within a predefined bounding box and filtering those with a predicted confidence score exceeding the threshold $\lambda$: $P_{rec}=\{ q|f_\theta\left({\widetilde{P}}_{o}';q\right)>\lambda\ \mathrm{or}\ f_\theta\left(P_{h_o};q\right)>\lambda\}$. Fig.\ref{Fig5} demonstrates the effectiveness of the pre-training component in point cloud reconstruction tests in both simulated and real-world environments.

Experiments conducted within the simulation environment demonstrate that the proposed method can accurately reconstruct the geometric shape and spatial pose of objects from hand-object interaction point cloud, even when the original point cloud is subject to the addition of noise and 70\% region masking. The L2 Chamfer Distance($CD-\ell_2$) is adopted as the similarity metric to evaluate the discrepancy between the reconstructed point cloud and the ground-truth point cloud of the target objects. After normalizing the hand-object interaction point cloud to a standardized space ${\widetilde{P}}_{o}' \in [-1, 1]^3$, the average reconstruction accuracy reaches a $CD-\ell_2$ of $8.1\times{10}^{-3}$, demonstrating the robustness of the model against perceptually ambiguous inputs.

In zero-shot transfer experiments conducted in real-world environments, with the incorporation of positional information from tactile sensors as supplementary input, the model effectively overcame challenges such as uneven noise distribution and sparsity of point clouds inherent in real scenes. The model successfully reconstructed point cloud structures that closely matched the shapes of the target transparent objects. These results demonstrate that the pre-trained encoder can effectively extract features representing the shape and pose of the interactive object even under ambiguous real-world scenarios, enabling it to provide reliable guidance for the downstream motor policy.

To quantify the contribution of the pre-training technique to the downstream motor policy, we compared the manipulation performance of the policy incorporating the pre-trained encoder against that of a policy trained from scratch. As shown in Table \ref{table2}, under identical experimental conditions, the pre-trained policy achieved an average success rate of 60\% higher than the policy trained from scratch. This indicates that pre-training significantly enhances the ability of the encoder to extract geometric features, thereby improving the robustness of the downstream motor policy in perceptually ambiguous scenarios.

\subsection{Ablation Study}
\begin{table}[t]
  \caption{Success Rate of Ablation Study}
  \label{table2}
  \centering
  \begin{tabular}{l c c c c}
    \toprule
    \textbf{Method} & \textbf{Pouring} & \textbf{Shaking} & \textbf{Rotating} & \textbf{Avg} \\
    \midrule
    Ours w/o Diff. Pred. & 0\% & 0\% & 0\% & 0\% \\
    Ours w/o Tac & 0\% & 0\% & 20\% & 7\% \\
    Ours w/o Pre-train & 30\% & 30\% & 10\% & 23\% \\
    \textbf{TransDex(Ours)} & \textbf{100\%} & \textbf{80\%} & \textbf{70\%} & \textbf{83\%} \\
    \bottomrule
  \end{tabular}
\end{table}
Ablation experiments were designed and conducted to analyze the impact of other components and information in the policy on manipulation performance. We evaluated the performance of three models with distinct ablations: one without differentiated prediction, one without tactile information, and one without pre-training. The results are presented in Table \ref{table2}. These indicate that differentiated prediction, tactile information and pre-training are all core components of the proposed policy. They contribute to success rate improvements of 83\%, 76\%, and 60\% respectively on transparent object manipulation tasks. Without tactile information, the policy failed to grasp the object in the pouring task. During the shaking task, the lack of force feedback often prevented the thumb and index finger from maintaining a stable pinch grip after the dexterous hand lifted the flask, leading to frequent drops of the flask. Notably, in contact-rich task, such as rotating glass cup, the policy without tactile information did not wait for finger contact with the cup and tactile feedback before initiating rotation, unlike the tactile-enabled policy. While this approach resulted in a non-zero success probability, it also made the cup prone to being knocked over due to uncontrolled contact forces.

The policy lacking differentiated prediction failed in all attempted trials. In the pouring and shaking tasks, the dexterous hand could grasp the objects but was unable to move them to the correct positions or adjust them to the target poses. In the rotating task, the policy demonstrated poor arm-hand coordination, failing to adjust the end-effector and finger positions based on tactile feedback. This resulted in severe oscillation and an inability to complete the task. These results suggest that the absence of differentiated prediction prevented the policy from capturing the correct synergistic arm-hand movement patterns. By contrast, incorporating feature combination and differentiated prediction significantly improves the success rate of complex, coordinated manipulation tasks involving the robotic arm and dexterous hand.

\subsection{Generalization to Complex Visual Interference and Unseen Objects}
To validate the generalization capability of TransDex, experiments were conducted to evaluate its performance in scenarios involving complex backgrounds and challenging lighting conditions, as well as in manipulating unseen objects, as illustrated in Fig.\ref{Fig4}. The experimental results are presented in Table \ref{table3}.

\begin{table}[t!]
  \caption{Success Rate of Generalization Study}
  \label{table3}
  \centering
  \begin{tabular}{l c c c c}
    \toprule
    \textbf{Settings} & \textbf{Pouring} & \textbf{Shaking} & \textbf{Rotating} & \textbf{Avg} \\
    \midrule
    Complex Backgrounds & 90\% & 80\% & 70\% & 80\% \\
    Complex Lighting & 80\% & 80\% & 60\% & 73\% \\
    Unseen Objects & 100\% & 90\% & 50\% & 80\% \\
    \bottomrule
  \end{tabular}
\end{table}

To validate performance under complex perceptual disturbances, experiments were conducted under two types of unseen complex backgrounds and one type of challenging lighting condition. The policy trained in the simple background was deployed directly without modification, with five attempts conducted for each scenario. The results demonstrate that TransDex achieved average success rates of 80\% and 73\% in scenarios involving complex backgrounds and challenging lighting condition, respectively. These results indicate that the 3D point cloud-based policy effectively resists interference from variations in background color and illumination, maintaining stable performance in transparent object manipulation tasks even under complex perceptual conditions. 

For each of the three tasks, two types of transparent objects that had not been seen before were selected as test objects, with five attempts conducted for each one. The results demonstrate that TransDex achieved an average success rate of 80\% on unseen objects. This indicates that the perceptual pre-training approach enhances performance by enabling the model to generalize the manipulation logic learned during training to objects outside the training dataset. Specifically, the policy achieved a 100\% success rate with both unseen objects in the pouring task. In the shaking task, all attempts with the narrow-mouth conical flask were successful, whereas one attempt with the wide-mouth flask failed. This failure occurred when the fingertips collided with the rim of the flask during the initial grasping attempt. For the rotating task, the success rate with unseen objects was 50\%, which is lower than the original success rate of 70\%. This is primarily because, after completing one rotation, the index and middle fingers often generated counteracting forces when re-establishing contact with the wider-aperture vase, which prevented sustained rotation. 

Furthermore, the performance of a policy based on 2D image information was compared in the rotating task under complex perceptual conditions. This policy adopted the original 3DTacDex architecture and retained its image encoder for processing 2D images. The experimental results show that with 3DTacDex, all attempts by the robot failed in the rotating task. This is primarily because the contours of transparent objects are easily confused with complex backgrounds in RGB images. By contrast, TransDex based on point clouds achieved a success rate of 70\% under the same conditions, performing significantly better than 3DTacDex. This validates the superiority of 3D point clouds over 2D images in capturing the geometric features of transparent objects. Robust perception and manipulation can be achieved through the spatial structure of the point clouds, supplemented by tactile information.

\section{Conclusion}
\label{sec:conclusion}
This paper presents TransDex, an end-to-end manipulation framework, integrating point cloud reconstruction pre-training and deep visuo-tactile fusion. TransDex demonstrates strong perceptual robustness and motion prediction capability in dexterous manipulation tasks of transparent objects. Experimental results show that the point cloud reconstruction pre-training technique allows the robot to accurately perceive target objects in perceptually ambiguous conditions, greatly improving the performance of the downstream policy in transparent objects manipulation. Furthermore, the framework improves the coordination and efficiency of the robotic system through feature combination and task adaptation. Notably, TransDex works independently of perceptual optimization modules, reducing potential error accumulation and latency in intermediate steps, and making it more suitable for dynamic, real-time robotic manipulation.

Nevertheless, this work has certain limitations. Firstly, the pre-training phase relies on simulated data. While the model exhibits a certain degree of transferability in real-world scenarios, it still requires further adaptation and adjustment for more complex environments. Secondly, the spatial resolution and response frequency of the tactile sensors restrict the performance in complex contact situations. Future work could involve constructing large-scale, real-world dexterous-hand interaction datasets; developing tactile-adaptive policies; and designing higher-performance tactile sensing systems. These developments would further enhance the framework's overall adaptability in dynamic and complex scenarios.

{\footnotesize
\bibliographystyle{IEEEtran}
\bibliography{references}
}

\end{document}